# Parallelization of Path Planning Algorithms for AUVs Concepts, Opportunities, and Program-Technical Implementation

Mike Eichhorn+
Institute for Automation and Systems Engineering
Ilmenau University of Technology
98684 Ilmenau, Germany
Email: mike.eichhorn@tu-ilmenau.de

Hans Christian Woithe, Ulrich Kremer*
Department of Computer Science
Rutgers University
Piscataway, New Jersey 08854
Email: {hcwoithe,uli}@cs.rutgers.edu

*Abstract* — **Modern autonomous underwater vehicles (AUVs) have advanced sensing capabilities including sonar, cameras, acoustic communication, and diverse bio-sensors. Instead of just sensing its environment and storing the data for post-mission inspection, an AUV could use the collected information to gain an understanding of its environment, and based on this understanding autonomously adapt its behavior to enhance the overall effectiveness of its mission. Many such tasks are highly computation intensive.**

**This paper presents the results of a case study that illustrates the effectiveness of an energy-aware, many-core computing architecture to perform on-board path planning within a battery-operated AUV. A previously published path planning algorithm was ported onto the SCC, an experimental 48 core single-chip system developed by Intel. The performance, power, and energy consumption of the application were measured for different numbers of cores and other system parameters. This case study shows that computation intensive tasks can be executed within an AUV that relies mainly on battery power. Future plans include the deployment and testing of an SCC system within a Teledyne Webb Research Slocum glider.**

## I. INTRODUCTION

Path planning algorithms for autonomous underwater vehicles (AUV) in a time varying environment require computationally intensive tasks like using ocean current information from a forecast system, accurate vehicle and cost models and the inclusion of inaccuracies. On the other hand, a fast calculation in mission planning is highly beneficial, especially when analyzing possible scenarios before mission start or during a mission, where a new route must be calculated with as little delay as possible using new information in the region of interest. These requirements can be met with parallel processing of the algorithms on multi-core processors. Such a multi-core processor used in an AUV could do the work of a whole control center on board, resulting in a significant improvement in the fields of application and the effectiveness of AUVs.

The Intel SCC (single-chip cloud computer) has been designed to implement a cloud data center in silicon on a single chip [1]. The research chip has 48 cores grouped in pairs of two cores (tiles), a 24 router-mesh on-chip network with 256 GB/s bisection bandwidth between tiles, and four integrated DDR3 memory controllers [2, 3]. Each core runs its own OS, thereby acting as an individual compute node. There is hardware support for message passing, but no hardware cache coherency policy is implemented.

The SCC system allows the power/energy management of individual cores and groups of cores, the on-chip network, and memory. Cores can be turned on and off. Frequency and voltage settings are software controlled and can be changed on the fly. This dynamic, fine grain power/energy management feature is the main characteristic of the SCC that we would like to exploit. The SCC is able to provide a significant range of energy vs. performance tradeoffs, giving AUVs the ability to perform mission critical, computation intensive tasks at the lowest possible energy cost. The SCC consumes between 25 and 125 Watts when all cores are active [4]. The speed of the on-chip network and off-chip memory can be adjusted, giving additional opportunities for performance vs. energy tradeoffs. The SCC is an experimental platform and not commercially available. As part of an ongoing collaboration with Intel, we have acquired an SCC system (shown in Figure 1) that we are evaluating for deployment within one of our Slocum gliders.

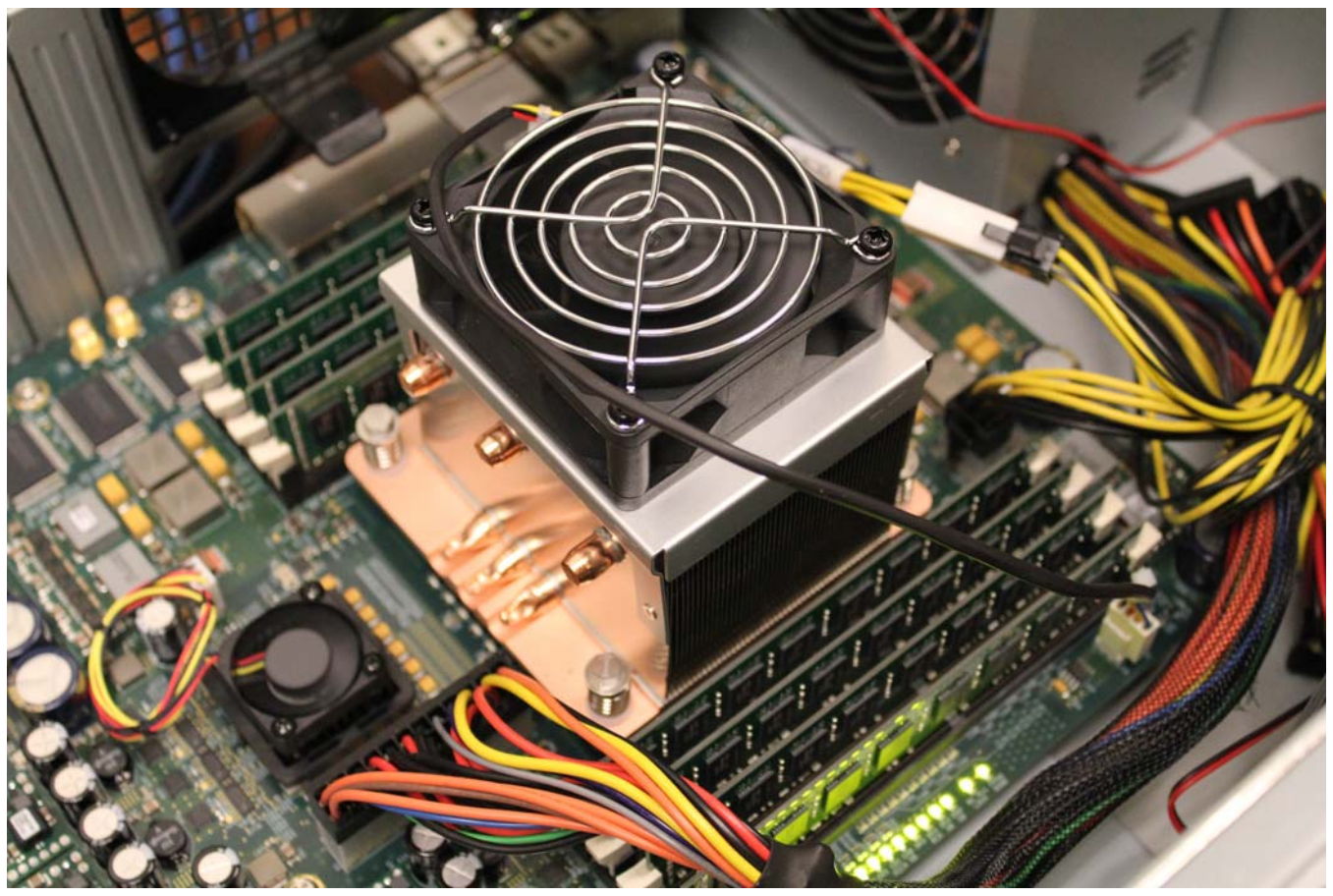

Figure 1. SCC System in our lab configuration

* This work was partially supported by NSF award CNS-MRI #0821607 and a hardware donation of an SCC system by Intel.
+ This work was supported by the Internal Excellence Promotion for Research at the Ilmenau University of Technology.

The effectiveness of dynamic frequency and voltage scaling is well known for dynamic power management. Switching between voltages/frequency settings and hibernation states incurs overhead costs, both in terms of performance and energy. Using these techniques in the context of high-performance systems and data centers has been executed with success [5, 6]; our work is novel because (1) it targets battery-operated environments and (2) it not only power/energy manages the CPUs but the on-chip network and the memory as well.

In the context of AUV applications where significant computations have to be performed under a soft real-time deadline, the parallelism available in the application and the SCC will be used to select a specific number of cores at a specific voltage/frequency level with specific network and memory speeds such that the desired deadline is met with the lowest energy consumption. In CMOS technology, power is proportional to the square of the supply voltage, and therefore frequency and supply voltage are essentially proportional ($P \sim f\, V^2_{supply}$). A strategy that performs the same work (operations) but on more, but slower components, can result in significant energy savings without violating execution deadlines.

Many sensor data processing and ocean modeling applications that are of interest to marine scientists contain substantial parallelism. These applications model or analyze physical phenomena in time and space, which maps well into parallel execution models. A prime example for parallel processing in AUVs is onboard path planning, and is the subject of this paper. Two possible scenarios for such an application are explained herein.

- Limited or unavailable communication to the base station

A glider has long operation periods up to 30 days. During this time the glider collects oceanographic data in the region of interest. Communication to a base station during this time is limited by the data bandwidth of satellite communication as well as the communication required energy and time. In the case of atmospheric disturbance, stormy weather or high seas communication is nearly impossible. In such cases, the glider must calculate its own route based on its knowledge of the ocean current behavior in the region of interest. Such information is implemented in tide tables, long term ocean current models and nautical experience. Compared to ocean current forecast information, these data are more inaccurate. An onboard robust path planning algorithm, presented in [7] can use such information to find a possible route. This approach requires parallel processing through multiple calculations of the parameter sets.

- Using collected ocean current data for planning

When a glider has an Acoustic Doppler Current Profiler (ADCP) onboard, the detection of actual ocean current information is possible. These data are much more accurate than any forecast information. The inclusion of collected ocean current information in a regional ocean modeling system can provide onboard forecast information for the region of interest. Both the forecast calculation and the following path planning calculation are computation intensive [8] and can benefit from a parallel execution.

In [7] possible concepts for the parallelization on multiple levels of the graph based path planning algorithms are discussed (see Figure 2). The necessity of data transfer in several levels correlates with their depths. In the lower levels the frequency of data transfer increases, which can result in poor performance with decreasing task granularities. Fine-grained tasks can negate the benefit of parallel execution due to the communication overhead.

Based on our experience with time computing cost functions using real ocean forecast data [9], the task of "Calculation of the optimal dive profile" (second block from right in Figure 2) was chosen as the first part of our path planning algorithm to be parallelized.

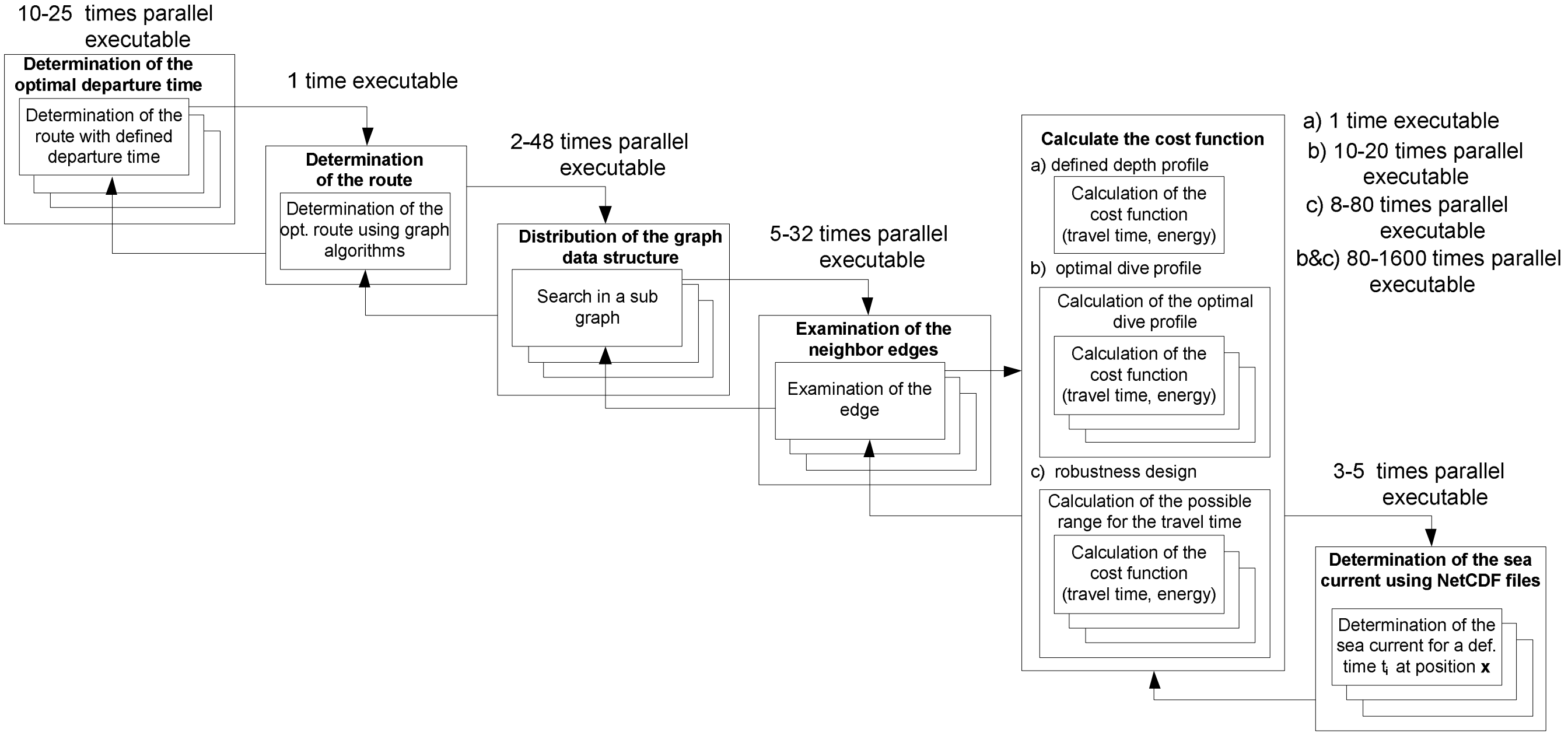

Figure 2. Opportunities to parallelize graph based path planning algorithms [7]

## II. PROGRAM TECHNICAL IMPLEMENTATION

### A. *Software Technical Details*

The path planning program is written in C++ using Microsoft Visual Studio 2010. This allows for easy programming and debugging with the MPI-Cluster debugger. To run the code on the SCC, the programs were compiled with GCC 3.4.5. Figure 3 shows the individual program parts and the libraries used (shaded dark grey) in overview. XML files are used to initialize the program with the mission parameters and to store the identified path after a search.

### B. *Message Passing Interface*

The serial version of our Time Varying Environment (TVE) path planning program (S-TVE) is a simplified and stripped down version of the path planning application presented in our previous work [9, 16, 17]. The parallel implementation (P-TVE), is based on the serial planning program and makes use of Message Passing Interface (MPI). MPI is a programming framework to help users write message-passing programs for parallel computers [10].

The P-TVE uses a master/slave architecture. The master performs the serial portions of the planning algorithm; it distributes work and collects results from the slaves. The slaves are tasked in parallel to calculate the costs associated for a glider to fly with various depth profiles.

Depending on the target platform, the MPI implementation used is either MPICH2 or Rock Creek MPI (RCKMPI) [11]. MPICH2 is well established and portable, while RCKMPI is SCC specific. The standard version of MPICH2 can also be used on the SCC if necessary. RCKMPI is itself based off of MPICH2 but takes advantage of the unique features of the SCC to decrease transmission time and increase program performance. For example, instead of using TCP/IP to communicate, RCKMPI can use the SCC's message passing buffer (MPB) or shared memory to transmit data among the nodes. The MPB is a small buffer on each of the SCC's tiles which allows for inter-tile communication within the chip. Generally, for small messages the MPB if preferred, while for larger messages, a shared memory mechanism is preferred. This distinction is due to the overhead required to break up and reassemble the larger messages to fit into the MPB [11].

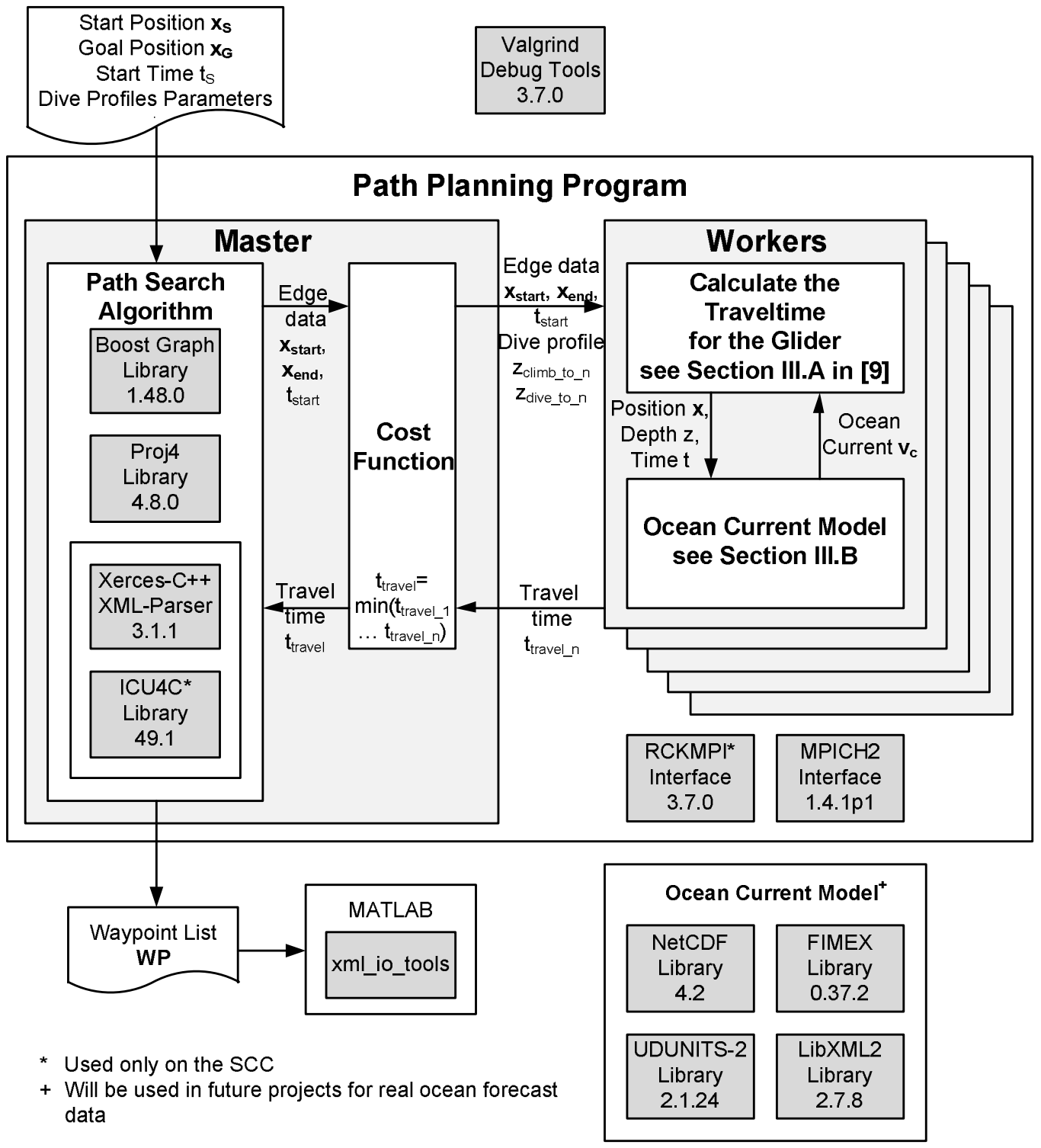

Figure 3. Software concept of the path planning program

Several SCC alternative message passing frameworks exist that could have been used instead of RCKMPI. RCCE, [4], is a message passing application programming interface (API) provided by Intel specifically for the SCC. The API allows low level access to the SCC including direct access to the MPB. The pipelined communication functions of the iRCCE extension to RCCE may provide performance improvements over RCCE [12]. Benchmarks in [11] demonstrate that RCCE and RCKMPI transmission overheads are comparable. The main advantage of using RCKMPI is that because it is based off of MPICH2, it uses the same API.

Using a standardized programming framework such as RCKMPI has the advantage that porting our path planning algorithm onto different target systems such as the SCC can be accomplished by linking in the appropriate library implementations, rather than making any changes to the source code itself, i.e., reprogramming the application. This also allows for easier testing of the program on other platforms. Coincidentally, the follow up to RCKMPI, invasive MPI (iMPI) [13], should provide improvements but was not as mature as its predecessor at the time we performed the study discussed in this paper.

To run the P-TVE program on the SCC, the user must initiate the program on an SCC core. The maximum number of possible parallel tasks is defined by the number of nodes in the message passing ring. This ring must be created before any parallel path planning can occur. Generally, this is performed once the SCC cores have been initialized and have booted Linux.

Before any execution can occur, MPI must first distribute the program to the nodes and perform any necessary bootstrapping. The zero-ranked node is designated the master while the others are slaves. After initialization, the slaves will immediately wait for instructions from the master using a blocking receive call. In RCKMPI, a blocking receive call performs continuous polling [11], which can lead to increased CPU utilization. Although this method decreases message latency, it increases the overall power of the SCC. This is undesirable for energy constrained, battery operated, devices like the glider. Therefore, one of the first tasks performed by the master is to instruct the slaves to enter sleep mode since the master node will need to perform several serial tasks before the parallel depth profile calculations are called.

Instead of blocking on a receive call, a slave's sleep mode is implemented as a non-blocking receive call that contains a sleep command while polling for messages. Slaves in this mode can still receive messages from the master. The tradeoff is lower CPU utilization, and therefore power, for increased message latency. The increased latency is only for the first task to be processed as the master sends a command to the required number of slaves to exit the sleep mode. The slaves will again continuously poll, with lower latency, waiting to receive tasks to be performed. After completion of the parallel portion of the program the master can again put the slaves in sleep mode until program termination or the slaves are needed again.

As we previously mentioned, slaves do not automatically enter sleep mode at program start. This design was initially chosen because it is the most flexible. It may be the case that the serial code at the beginning of the program is known to be short, and therefore it may not be worthwhile to put the slaves to sleep automatically. We leave this policy decision for future work because the most energy efficient mode for real deployments and real input data should be decided for each mission individually. The benefits that can be gained by the parallelization of the algorithm are affected by the input parameters to the program. In our implementation, we perform the optimal dive profile calculation simultaneously so that input parameters which favor more of these calculations will facilitate a greater speed increase compared to the serial version.

## III. RESULTS

This section presents the results of the path planning algorithm written in C++ and executing on the SCC. The first part of this section describes the chosen parallelization concept which is also presented in [9]. It is located in the lower part of the parallelization hierarchy of Figure 2. The parallelization processes occur in the cost function, called in the search algorithm (see Figure 3), to determine an optimal dive profile. We used ocean current models with a depth dependent ocean current to verify the search algorithm and its archived results will be discussed. Subsequently, a detailed analysis of the executed benchmarks using variable numbers of processor cores on the SCC will be described.

### *A. Parallelisation Concept -Detect Optimal Dive Profile*

Passing through regions with an adverse surface or seabed current by using a constant dive profile with possible large depth amplitude in order to collect oceanographic data at each depth, is impossible. The search algorithm will create a path to drive around these areas, taking the long way around to arrive at the goal point. Data in the region of interest will therefore not be collected. The simulation of a selection of dive profiles with different "climb-to" depths $z_{climb\text{-}to}$ and "dive-to" depths $z_{dive\text{-}up}$ distributed over the maximum permitted depth profile in every cost function can solve the problem [9].

The created dive profiles are specified through a minimum diving depth $z_{min}$, a maximum diving depth $z_{max}$, a maximum "climb-to" depth $z_{climb\text{-}to_max}$, a minimum dive amplitude $z_{minRange}$, and the number of "climb-to" $n_{climb\text{-}to_levels}$ levels and "dive-to" levels $n_{dive\text{-}to_levels}$. The pseudo code to determine the dive profiles *DP* is depicted in Table I. Figure 4 shows a possible dive profile selection. Every cost function calculates *n* travel time values for the various dive profiles according to the algorithm in section III.A in [9]. The profile with the least travel time provides the cost value that is used.

TABLE I
PSEUDO-CODE OF THE ALGORITHM TO CREATE THE DIVE PROFILES

DIVE-PROFILES ($z_{min}$, $z_{max}$, $z_{climb\text{-}to_max}$, $d_{minRange}$, $n_{climb\text{-}to_levels}$, $n_{dive\text{-}to_levels}$)
**if** ($n_{climb\text{-}to_levels} > 1$)
    $d_{climb\text{-}to_levels} = (z_{climb\text{-}to_max} - z_{min})/(n_{climb\text{-}to_levels} - 1)$
    $Z_{climb\text{-}to} = z_{min} : d_{climb\text{-}to_levels} : z_{climb\text{-}to_max}$
**else**
    $Z_{climb\text{-}to} = z_{min}$
**if** ($n_{dive\text{-}to_levels} > 1$)
    $d_{dive\text{-}to_levels} = (z_{max} - z_{min} - d_{minRange})/(n_{dive\text{-}to_levels} - 1)$
    $Z_{dive\text{-}to} = z_{max} : -d_{dive\text{-}to_levels} : (z_{min} - d_{minRange})$
**else**
    $Z_{dive\text{-}to} = z_{max}$
*k*=0
**for**(*i* = 1) **to** (*i* = length($Z_{climb\text{-}to}$))
    **for**(*j* = 1) **to** (*j* = length($Z_{dive\text{-}to}$))
        **if** (($Z_{dive\text{-}to}[j] - Z_{climb\text{-}to}[i]) > d_{minRange}$)
            $k = k + 1$
            $DP[k] = \{Z_{climb\text{-}to}[i], Z_{dive\text{-}to}[j]\}$
**return** *DP*

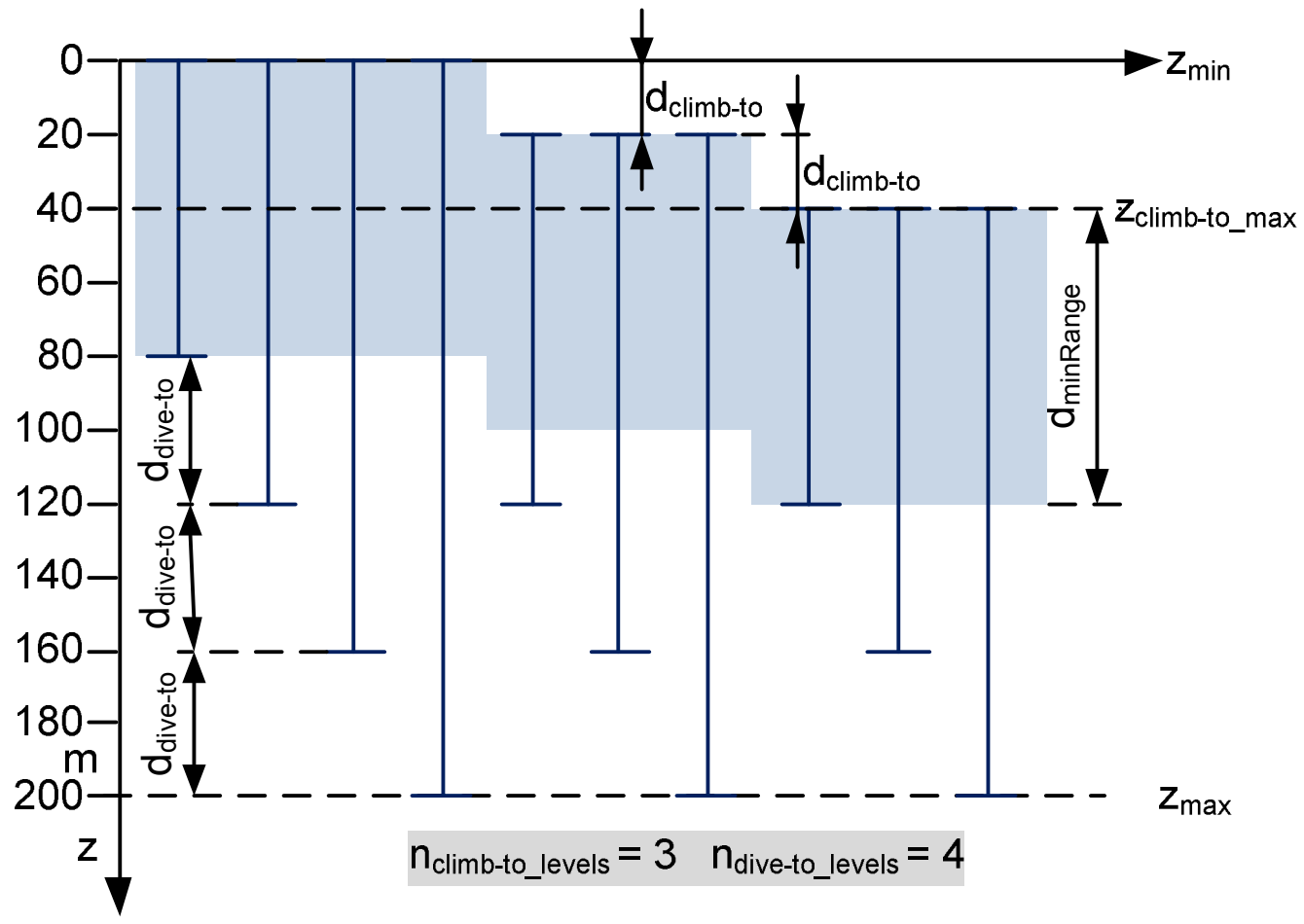

Figure 4. Created dive profiles [9]

### *B. The selected test function for a Time-Varying Ocean Flow*

The function used to represent a time-varying ocean flow describes a meandering jet in the eastward direction, and is a simple mathematical model of the Gulf Stream [14, 15]. This function was applied in [7, 9, and 16] to test the TVE algorithm and its modifications. To verify the path planning algorithm a depth dependent ocean current will be added to this model. The stream function is:

$$\phi(x,y) = 1 - \tanh\left(\frac{y - B(t)\cos\left(k\left(x - ct\right)\right)}{\left(1 + k^2 B(t)^2 \sin^2\left(k\left(x - ct\right)\right)\right)^{\frac{1}{2}}}\right) \quad (1)$$

which uses a dimensionless function of a time-dependent oscillation of the meander amplitude

$$B(t) = B_0 + \varepsilon\cos(\omega t + \theta) \quad (2)$$

and the parameter set $B_0 = 1.2$, $\varepsilon = 0.3$, $\omega = 0.4$, $\theta = \pi/2$, $k = 0.84$ and $c = 0.12$ to describe the velocity field:

$$u(x,y,t) = -\frac{\partial \phi}{\partial y} \quad v(x,y,t) = \frac{\partial \phi}{\partial x}. \tag{3}$$

To descript the depth dependence ocean current a time and depth variant term $u_{surface}(z,t)$

$$u_{surface}(z,t) = W(t)\max\left(\left(1 - \frac{1}{z_{max}} z\right), 0\right) \quad v(z,t) = 0 \tag{4}$$

$$W(t) = W_0 \cos(d\omega t) \tag{5}$$

with the parameters $z_{max} = 15$ m, $W_0 = 0.5$ and $d = 2$ will be add to the $u(x,y,t)$ component of Equation (3):

$$\begin{aligned} u(x,y,z,t) &= u(x,y,t) + u_{surface}(z,t) \\ v(x,y,z,t) &= v(x,y,t) \end{aligned}. \tag{6}$$

This additional term shall describe the influence of the wind on the surface current. The influence decreased linear up to a depth $z_{max}$ as well as being time variant with an angular frequency $d\omega$ (see (5)). As a result $u_{surface}$ emulates the behaviour of a head- (negative values for $W(t)$) and tailwind (positive values for $W(t)$) in x-direction. Figure 5 shows the resultant ocean current field with the influence of $u_{surface}$ at different times. Negative $u_{surface}$ values lead to surface eddies, which should be avoided by the AUV. The dimensionless value for the body-fixed vehicle velocity $v_{veh_bf}$ is 0.5.

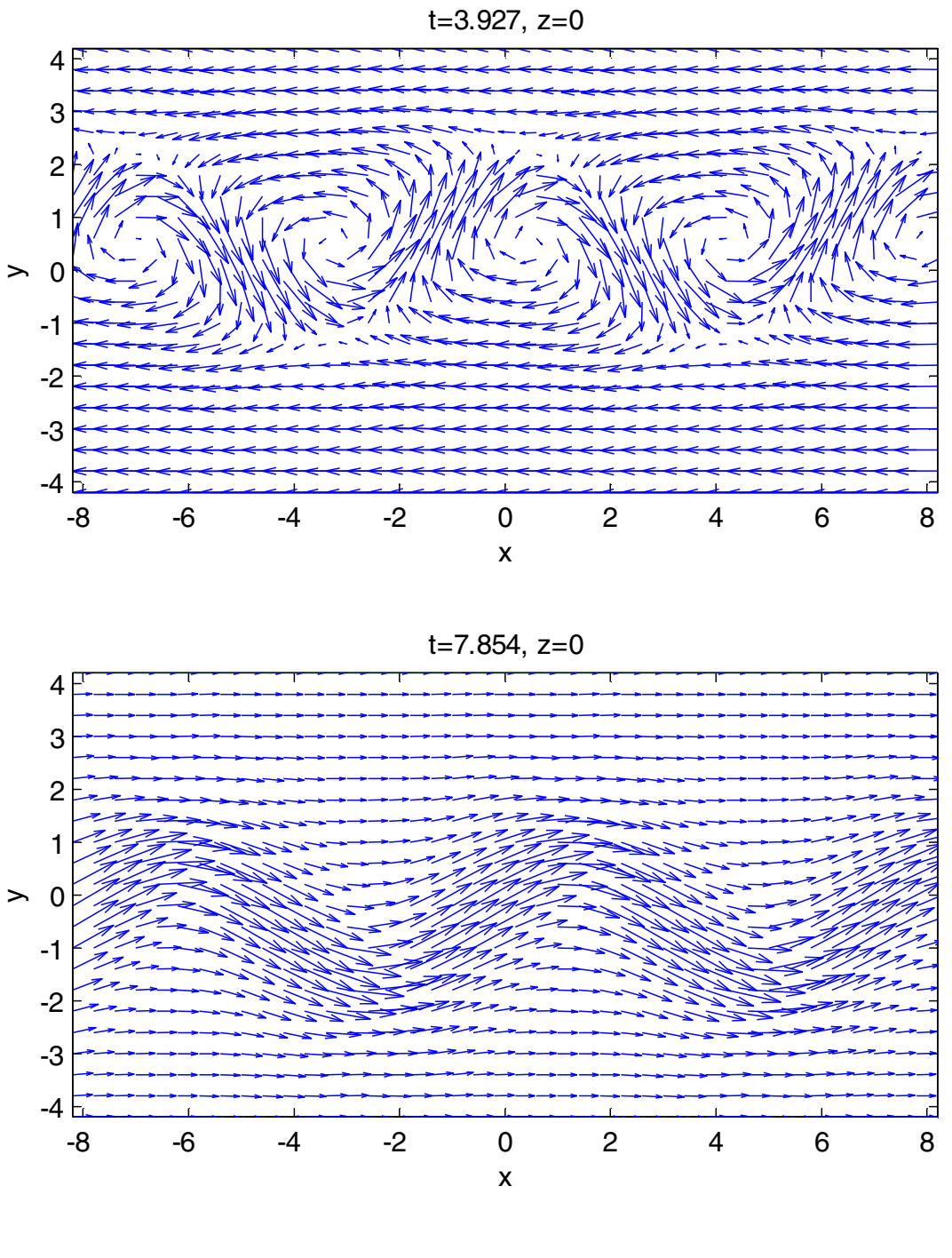

Figure 5. Ocean current field on the surface by head- (upper figure) and tailwind (lower figure)

## C. Test conditions and results

For the benchmark tests on the SCC, the rectangular 3-sector grid structure with a grid size of 0.4 was used ([17]). Several dive profiles were created from the calculations in Table I with the parameters $z_{min} = 0$ m, $z_{max} = 200$ m, $z_{climb\text{-}to_max} = 40$ m, $z_{minRange} = 50$ m, $n_{climb\text{-}to_levels} = 4$ and $n_{dive\text{-}to_levels} = 6$. These parameters were chosen to conform to glider behavior and the provided ocean current information from a forecasting system of a real mission. This results in 20 dive profiles. The value of the parameter $z_{climb\text{-}to_max}$ was chosen for safety as a larger value could cause the buoyancy pump to draw in too quickly. The chosen value for the parameter $z_{minRange}$ enables a stable flight behavior. The resolution of the climb-to and dive-to levels is defined by the current data from a forecast system which provides data for several depth layers.

For the tests, the Time Varying Environment (TVE) algorithm was used. It is presented in section II.A in [16]. Figure 6 shows the time sequence of the course through the time-varying ocean flow identified by the TVE algorithm. The solution has the characteristic that the vehicle drives in the main-stream of the jet and avoids eddies. The dive behaviour of the vehicle in the case of a counter-current during a mission can be clearly seen in Figure 7. The vehicle ascends only until 26.66 m to avoid the adverse ocean current on the surface.

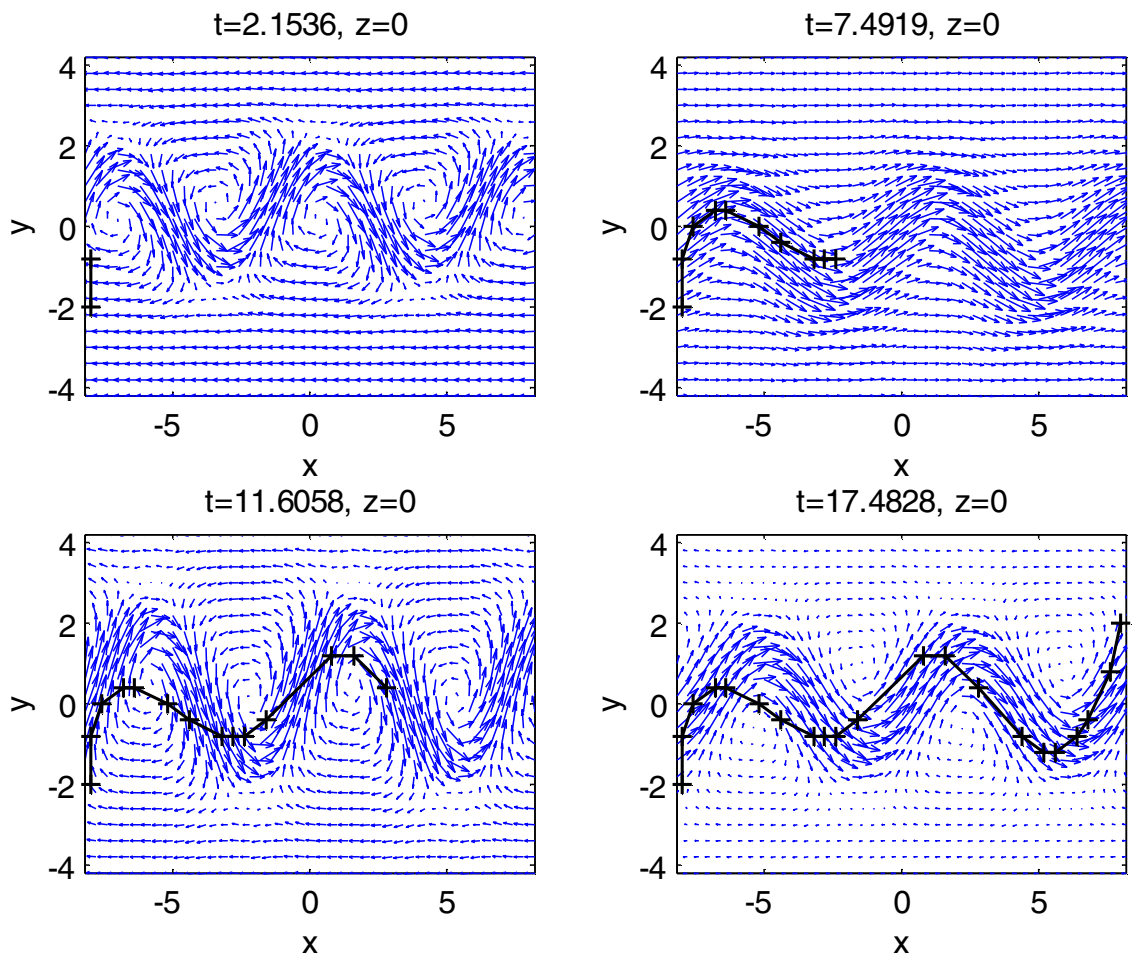

Figure 6. Time sequence of the predicted paths through the time–varying ocean current field by the TVE algorithm with a 3 sector grid sructure

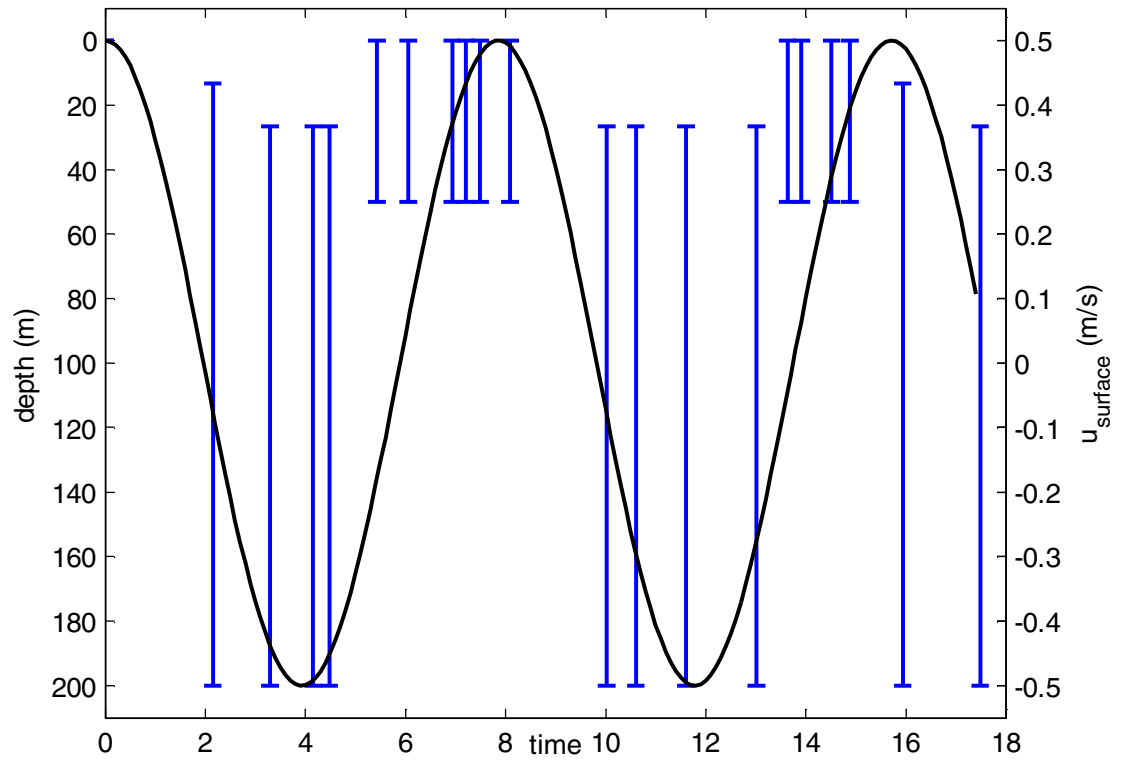

Figure 7. Predicted dive profils for the several path elements

*D. Benchmarks*

We have evaluated serial TVE (S-TVE) and parallel TVE (P-TVE) on the SCC. The SCC was initialized to run with the highest default setting profile provided by Intel. The cores in this profile run at 800MHz, the mesh network at 1600MHz and the DDR memory controller at 1066MHz. The input parameters for both versions of the program were the same and were identical to the ones described in the previous section.

The runtime performance results from our evaluation are shown in Figure 8. Since there is no parallelism involved in S-TVE, it is only executed on a single core. P-TVE requires at least two cores, a master and a slave; therefore no results exist for a single core. The number of slaves is always one less than the number of cores. In the case of 48 cores, for example, 47 cores are slaves and can perform the dive profile calculation.

The MPI-NOOP results measure the overhead of the MPI infrastructure used in our experiments. MPI-NOOP is a modified version of P-TVE but performs no actual path planning. The program simply runs, initializes MPI, and immediately exits. For two cores, P-TVE performs worse, by having a longer runtime, than S-TVE. This is due to the communication overhead from master to slave which does not exist in S-TVE. However, without this exception, P-TVE outperforms S-TVE because the speedup gained in the simultaneous depth profile calculations pay off.

In Figure 9, the runtime of only the depth profile search calculation is shown. This search runtime does not include the serial portions of the algorithm including the construction of graphs. In conjunction with Figure 8, there is a cost associated with running the algorithm on only two cores. The step-wise degradation of runtime is more easily observed and is especially noticeable in Figure 10 when graphing the speedup of the search calculation. The speedup is normalized to the path search time of S-TVE and is therefore zero in the figure for S-TVE. In this evaluation, we explore a maximum of 20 depth profile calculations for each edge in the graph. Since we only explore a given edge at a time, in this implementation of P-TVE we gain no benefit by using more than 21 cores (20 slaves). This accounts for the lack of speedup and no decrease in runtime after 21 cores. If the number of depth profiles were increased, additional cores could be used with a concomitant increase in benefit.

The step-wise behavior of speedup/runtime for less than 21 cores can be explained by the number of work delegation iterations that are required to be distributed to the slaves by the master. For example, for both six and seven cores (five and six slaves respectively), four rounds of delegation are required. Furthermore, with seven cores, the fourth round will only contain two depth profiles to calculate, leaving four slaves idle. For eight nodes (seven slaves), however, we require only three rounds with the final round having six depth profiles. Eleven nodes require two rounds. Finally, for 21 nodes, only one iteration of work delegation is needed since there is one slave for each depth profile calculation. A decrease in the speedup can also be observed in Figure 10 after each step. This is especially noticeable after 21 cores. Although unused cores go into a sleep mode, they still poll periodically until the path search is complete.

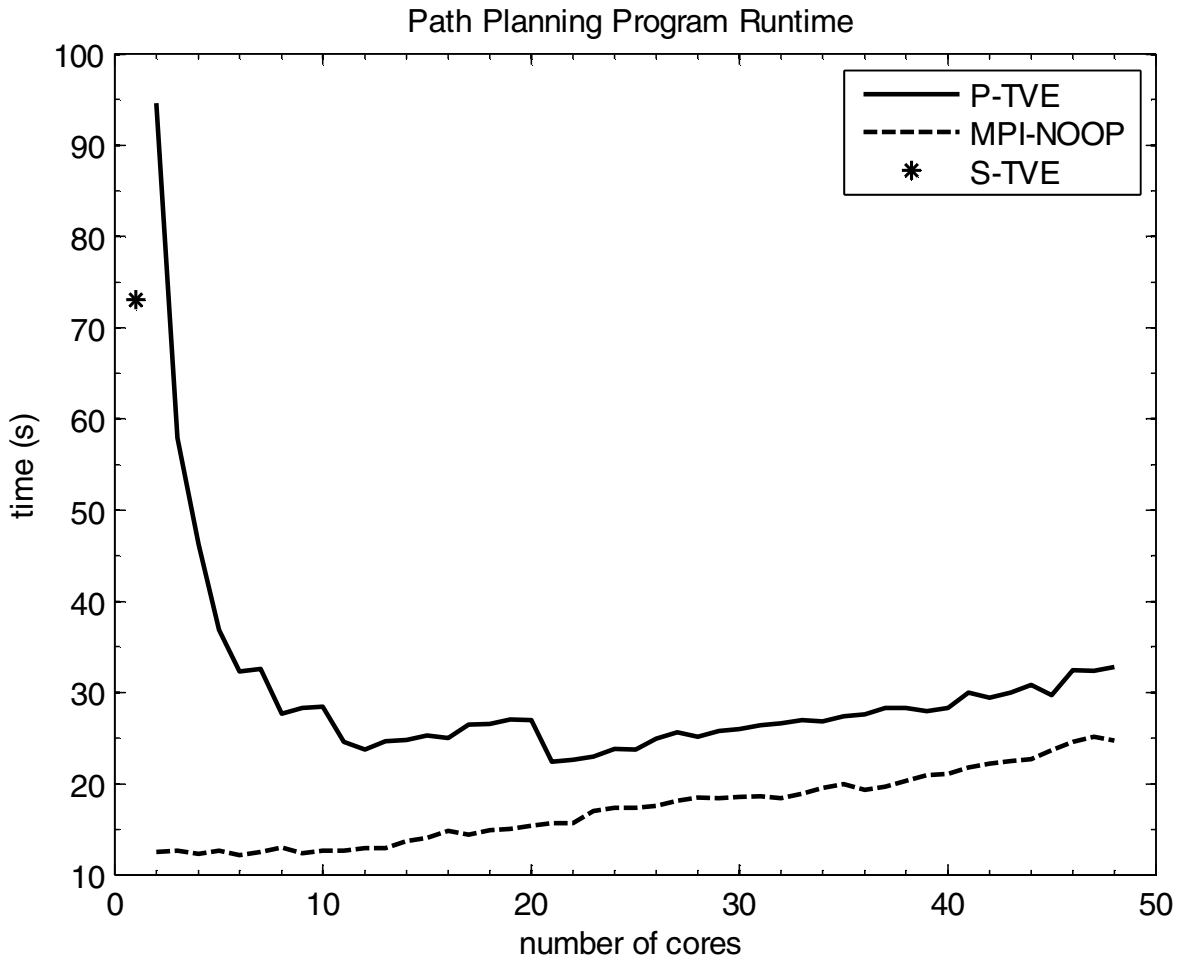

Figure 8. Runtime of the path planning and MPI-NOOP programs

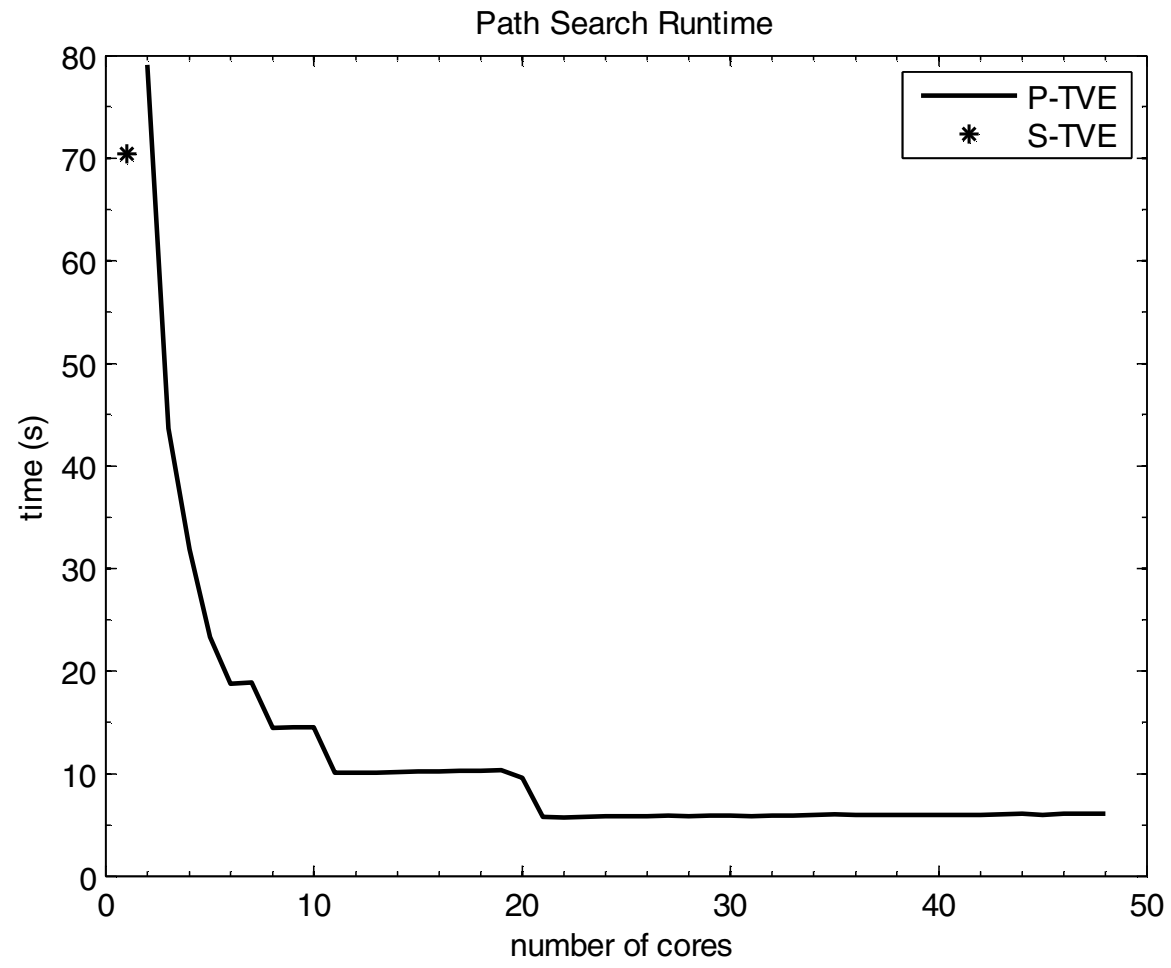

Figure 9. Runtime of the seach algorithm of the serial and parallel programs

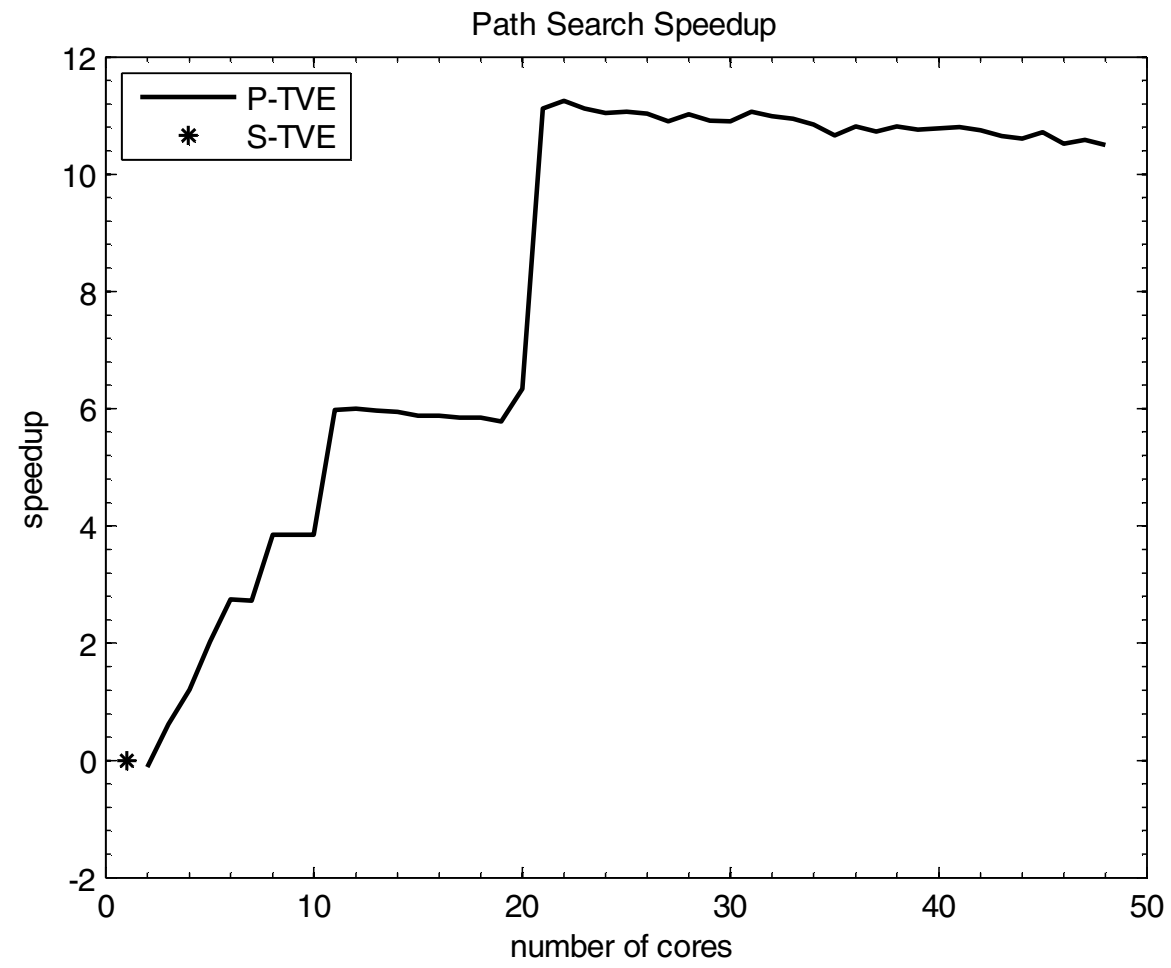

Figure 10. Speedup of the seach algorithm normalized to the serial program

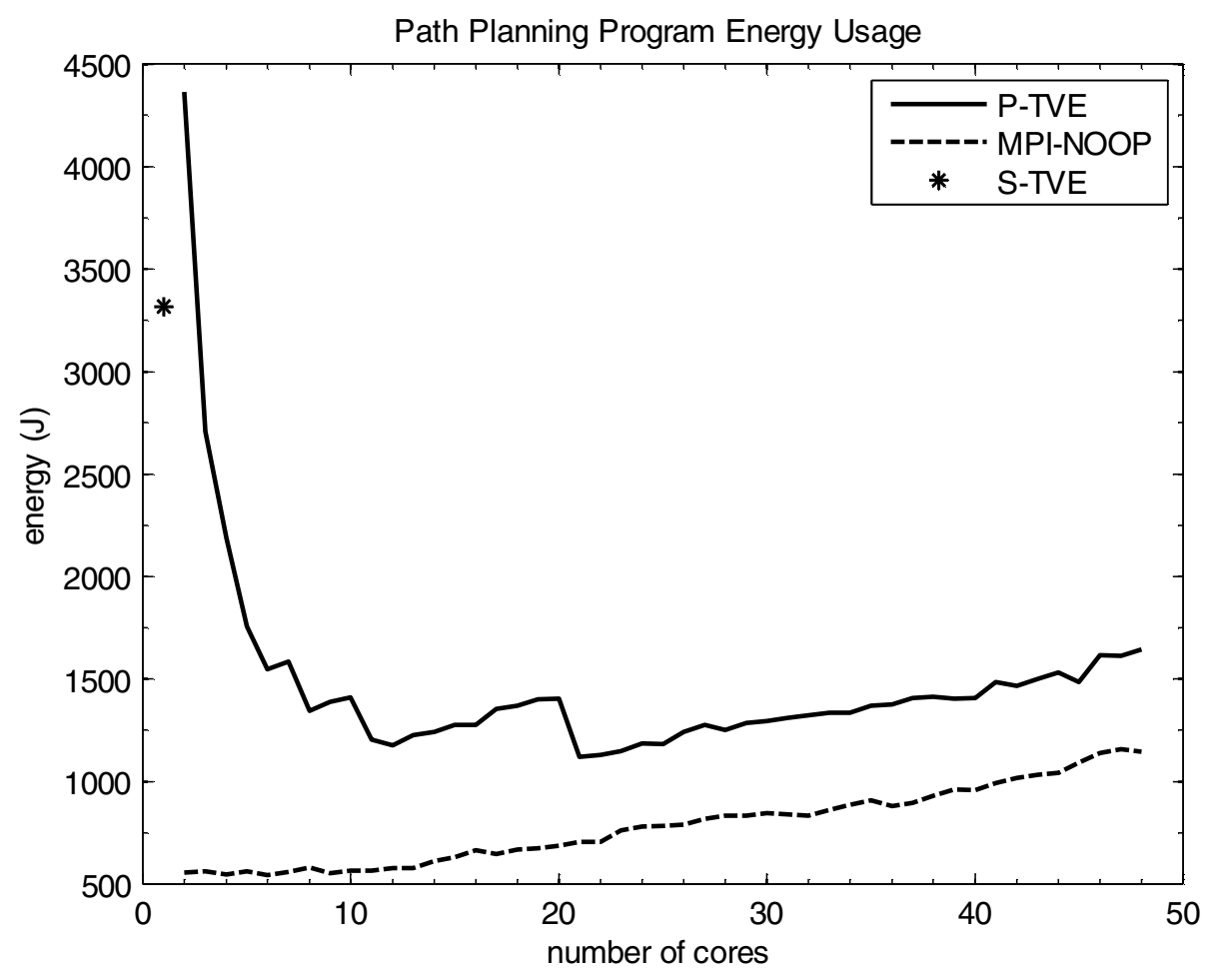

Figure 11. Energy consumption of the path planning and MPI-NOOP programs

The sleep time of the polling receive used in this evaluation was 100 ms, and could be the cause of increased message latency. The runtime difference of the parallel dive profile search between 21 and 48 cores is approximately 300 ms. Although minor, this can be avoided by preventing the use of cores that were not needed in the first place.

The cost of increasing the number of cores past 20 slaves does not greatly affect the profile search runtime/speedup, but does have a significant impact on the runtime of the entire program execution. In our experiments, we use the Multiple Purpose Daemon (MPD) process manager from MPICH2 to create and manage the MPI ring. MPD uses SSH over TCP/IP to launch the master and slave nodes. This causes the launch time to increase with the number of cores and can be observed in MPI-NOOP runtime in Figure 8. From this, we can conclude that this increase is in the overhead of the infrastructure itself and not P-TVE. Similar effects have been observed and are addressed in [13]. The process manager (PM) of iMPI should mitigate the program launch times. As part of future work, we plan to migrate to this MPI alternative.

The energy required to execute the path planning programs is shown in Figure 11. The general trend is similar to the runtimes in Figure 8. By utilizing more cores, P-TVE is capable of saving significant energy over S-TVE. Even within P-TVE, using 21 cores instead of 11 uses less energy and has a lower runtime. Higher numbers of cores incur the penalty of increased launch times, even if the cores are unused, and therefore are not as energy efficient. The PM of iMPI should nearly eliminate this overhead. Thus, if most of the MPI-NOOP energy and runtime could be subtracted from the P-TVE, the gains of parallelizing the path planning algorithm would be substantial.

## IV. CONCLUTIONS AND FUTURE WORK

In this paper we have described several opportunities for the parallelization of a path planning algorithm. We have also implemented and evaluated one of these opportunities by parallelization of the dive profile calculations among nodes in a single chip computing cluster. In the master/slave architecture employed, the master node computes the serial portions of the algorithm and distributes the computing profile tasks to slaves.

Evaluation of the parallel path planning approach indicates that performance improvements can indeed be made. For the chosen input parameters, the parallel version generally outperforms the serial version both in runtime and in energy. If the number of possible dive profiles were to be increased, then the parallel program would gain further speedup. Benchmarks of a simple MPI program suggest that the overhead of launching and initializing the programs on the nodes can lead to a decrease in runtime and energy performance if nodes have no computation to perform and are idle. This overhead, based on related work, suggests that this overhead can be mitigated by using an MPI process manager that takes advantage of the SCC's unique features.

The presented results indicate that state-of-the-art parallel platforms such as the SCC can enable the execution of computation intensive tasks in energy constrained environments such as an AUV. Marine scientists can take advantage of this novel capability and make AUVs more effective and autonomous research platforms. We are in the process of porting a ROMS ocean modeling code onto the SCC as well. We are also planning to install an SCC and a multi-processor ARM cluster within one of our Slocum gliders, and perform field tests in the Atlantic Ocean off the coast of New Jersey.

## V. ACKNOWLEDGEMENT

We would like to thank Intel for providing the EEL lab at Rutgers with an SCC system in support of this research. We are particularly grateful to the SCC development team for their help and advice. The authors would like to thank William Brozas and Bharath Pichai for their support during this project.

## REFERENCES

[1] Intel Research, "Single-Chip Cloud Computer," 2012, http://techresearch.intel.com/ProjectDetails.aspx?Id=1.

[2] Gries, M., Hoffmann U. Konow, M. and Riepen M., J. O. - Computing in Science Engineering, "SCC: A Flexible Architecture for Many-Core Platform Research," Computing in Science Engineering, vol. 13, NU - 6, no. 6, 2011, pp. 79 -83

[3] Howard, J. et al., "A 48-Core IA-32 Processor in 45 nm CMOS Using On-Die Message-Passing and DVFS for Performance and Power Scaling," Solid-State Circuits, IEEE Journal of, vol. 46, no. 1, 2011, pp. 173 -183.

[4] Mattson, Timothy G. et al., "The 48-core SCC Processor: the Programmer's View," in Proceedings of the 2010 ACM/IEEE International Conference for High Performance Computing, Networking, Storage and Analysis, IEEE Computer Society, 2010, pp. 1-11.

[5] C. Hsu, W. Feng, "A Power-Aware Run-Time System for High-Performance Computing," SC '05 SC Conference on High Performance Computing Networking, Storage and Analysis, 2005.

[6] T. Heath, B. Diniz, E.V. Carrera, W. Meira and R. Bianchini, "Energy conservation in heterogeneous server clusters," PPoPP '05 Proceedings of the tenth ACM SIGPLAN symposium on Principles and practice of parallel programming, 2005.

[7] M. Eichhorn and U. Kremer, "Opportunities to Parallelize Path Planning Algorithms for Autonomous Underwater Vehicles," Oceans '11 IEEE Kona, 2011.

[8] H.C. Woithe, I. Chigirev, D. Aragon, M. Iqbal, Y. Shames, S. Glenn, O. Schofield, I. Seskar and U. Kremer, "Slocum Glider Energy Measurement and Simulation Infrastructure," Oceans '10 IEEE Sydney, 2010.

[9] M. Eichhorn, C. Williams, R. Bachmayer and B.d. Young, "A Mission Planning System for the AUV “SLOCUM Glider” for the Newfoundland and Labrador Shelf," Oceans '10 IEEE Sydney, 2010.

[10] MPICH2 “MPICH2 website”, 2012, http://www.mcs.anl.gov/research/projects/mpich2/

[11] Ureña, Isaías A. Comprés, Riepen, Michael and Konow, Michael, RCKMPI - lightweight MPI implementation for intel's single-chip cloud computer (SCC), Springer-Verlag, 2011

[12] Clauss, C., Lankes, S., Reble, P. and Bemmerl, T., "Evaluation and improvements of programming models for the Intel SCC many-core processor," in High Performance Computing and Simulation (HPCS), 2011 International Conference on, 2011, pp. 525 -532.

[13] Ureña, Isaías, Riepen, Michael, Konow, Michael and Gerndt, Michael, "Invasive MPI on Intel’s Single-Chip Cloud Computer," in Architecture of Computing Systems – ARCS 2012, vol. 7179, A. Herkersdorf, K. Römer and U. Brinkschulte, Edt., Springer Berlin / Heidelberg, 2012, pp. 74-85.

[14] M. Cencini, G. Lacorata, A. Vulpiani and E. Zambianchi, "Mixing in a Meandering Jet: A Markovian Approximation," Journal of Physical Oceanography, vol. 29, 1999, pp. 2578-2594.

[15] A. Alvarez, A. Caiti and R. Onken, "Evolutionary Path Planning for Autonomous Underwater Vehicles in a Variable Ocean," IEEE Journal of Oceanic Engineering, vol. 29, no. 2, 2004, pp. 418-428.

[16] M. Eichhorn, "Solutions for Practice-oriented Requirements for Optimal Path Planning for the AUV “SLOCUM Glider”, Oceans '10 IEEE Seattle, 2010.

[17] M. Eichhorn, "A New Concept for an Obstacle Avoidance System for the AUV “SLOCUM Glider” Operation under Ice", Oceans '09 IEEE Bremen, 2009.